\documentclass[lettersize,journal]{IEEEtran}
\usepackage{amsmath,amssymb,amsfonts}
\usepackage{algorithmic}
\usepackage{algorithm}
\usepackage{array}
\usepackage[caption=false,font=normalsize,labelfont=sf,textfont=sf]{subfig}
\usepackage{textcomp}
\usepackage{stfloats}
\usepackage{url}
\usepackage{verbatim}
\usepackage{graphicx}
\usepackage{cite}
\usepackage{dsfont}
\usepackage{xcolor}
\begin{document}

\title{Hierarchical Multi-agent Meta-Reinforcement Learning for Cross-channel  Bidding}

\author{Shenghong He,
 Chao Yu*,
 Qian Lin, Shangqin Mao, Bo Tang,
 \\
 Qianlong Xie, Xingxing Wang
\thanks{Shenghong He, Chao Yu and Qian Lin are with the School of Computer Science and Engineering, Sun Yat-sen University, Guangzhou 510006, China (e-mail: yuchao3@mail.sysu.edu.cn)}
\thanks{
Bo Tang is with  the School of
Artificial Intelligence and Data Science, University
of Science and Technology of China. Hefei 340101, China
}

\thanks{Shangqin Mao, Qianlong Xie, and Xingxing Wang are with Meituan, Beijing 100102, China.}
}

\markboth{Journal of \LaTeX\ Class Files,~Vol.~14, No.~8, August~2021}%
{Shell \MakeLowercase{\textit{et al.}}: A Sample Article Using IEEEtran.cls for IEEE Journals}


\maketitle

\begin{abstract}
Real-time bidding (RTB) plays a pivotal role in online advertising ecosystems. Advertisers employ strategic bidding to optimize their advertising impact while adhering to various financial constraints, such as the return-on-investment (ROI) and cost-per-click (CPC). Primarily focusing on bidding with fixed budget constraints, traditional approaches cannot effectively manage the dynamic budget allocation problem where the goal is to achieve global optimization of bidding performance across multiple channels with a shared budget. 
In this paper, we propose a hierarchical multi-agent reinforcement learning framework for multi-channel bidding optimization. 
In this framework, the top-level strategy applies a CPC constrained diffusion model to dynamically allocate budgets among the channels according to their distinct features and complex interdependencies, while the bottom-level strategy adopts a state-action decoupled actor-critic method to address the problem of extrapolation errors in offline learning caused by out-of-distribution actions and a context-based meta-channel knowledge learning method to improve the state representation capability of the policy based on the shared knowledge among different channels. 
Comprehensive experiments conducted on a large scale real-world industrial dataset from the Meituan ad bidding platform demonstrate that our method achieves a state-of-the-art performance.

\end{abstract}


\section{Introduction}
Real-Time Bidding (RTB)~\cite{DBLP:GoldfarbT11,DBLP:WangZY17} is a programmatic advertising mechanism where advertisers participate in bidding through automated platforms when a user visits a webpage, securing the opportunity to display ads on the page through either a first-price auction~\cite{DBLP:despotakis2021first} or a second-price auction~\cite{DBLP:00020Z13}.
In RTB, advertising platforms formulate a bid request for every individual impression (i.e., display page) in real-time, and then advertisers submit bids for this impression utilizing their bidding algorithms. 
The bidding algorithms commonly assist in bid determination by considering factors in the users' preferences, requirement information, and other relevant issues, aiming to improve the revenue of the advertising platform (typically linked to user clicks) while at the same time adhering to constraints such as budget, cost-per-click (CPC) and return on investment (ROI), where CPC reflects the cost per ad click, and ROI is a measure of return on advertising investment.

An increasing number of advertising platforms have introduced automated bidding services in various ad settings, including recommendation ads (i.e., ads suggesting products to potential users) \cite{DBLP:GaoLCLZZ23, DBLP:ZhangWYLW23},  search ads (i.e., ads presented in response to user search queries)~\cite{DBLP:YuYXCLH24,koenecke2023popular}, and others~\cite{padghan2023aggregator, DBLP:YuYXCLH24}.
These ad settings typically involve multiple advertising channels, each of which corresponds to a specific medium or user-customized service for ad delivery.
However, most existing studies only focus on\textbf{ single-channel settings} where the bidding strategy is applied to a single channel to improve bidding performance therein~\cite{DBLP:CaiRZMWYG17, DBLP:HeCWPTYXZ21, DBLP:WangDFY0WZ22}. 
Unlike single-channel bidding, in cross-channel bidding, advertisers can bid on multiple advertising channels at the same time, which requires the bidding algorithm to take into account diversities in user characteristics and behaviors of different channels, and calls for more complex budget allocation and management strategies to ensure efficient and effective budget utilization across different channels.
Some studies~\cite{DBLP:WenXZZWLRXTYXWC22, DBLP:WangTLMZDSXWW24} have attempted to address the cross-channel bidding problem using single-channel methods. 
However, they have not considered the issues of channel budget allocation or interconnections between channels, which are critical in determining the final performance.
In this paper, we focus on \textbf{cross-channel bidding settings} where the goal is to achieve global optimization of bidding performance across multiple channels with a shared budget (i.e., the objective is to maximize the benefits for advertisers and advertising platforms while enhancing the user's overall client experience).

In this setting, there exist two main challenges to be addressed.
The first is  \textbf{the dynamic allocation of budgets among different bidding channels}.
A straightforward solution to this challenge is to share the total budget as a whole across different channels or allocate budgets to each channel based on a preset ratio.
However, as the optimal pairing between channels and advertisers can evolve over time, such a static allocation scheme may yield myopic budget allocations and thus lead to suboptimal performance.
Therefore, a better solution is to allocate the total budget dynamically to different channels based on the real-time information of channels and markets. 
The second challenge is \textbf{the utilization of cross-channel interrelationships for efficient bidding decision making within each channel}.
In the cross-channel setting, an effective bidding strategy should not only consider the characteristic of each channel but also incorporate bidding information from other channels to achieve global maximization of bidding performance. 
By harnessing their inherent relationships and knowledge sharing across various channels, it is capable of learning more efficient bidding strategies, particularly for those channels with only low customer traffic and thus limit data for training.

To tackle the above challenges, we propose a novel reinforcement learning (RL) approach called \textbf{H}ierarchical \textbf{M}ulti-agent \textbf{M}eta-reinforcement Learning for \textbf{C}ross-channel \textbf{B}idding (HMMCB) to solving the cross-channel bidding problem under a hierarchical RL framework. 
In HMMCB, the top-level strategy focuses on dynamically allocating budgets to each channel, while the bottom-level strategy centers on making bidding decisions by utilizing cross-channel interrelationships under the given budget constraint.
In specific, \textbf{at the top level}, budget allocation to multiple channels is determined by solving an optimization problem that maximizes user clicks at the bottom level while satisfying the CPC constraint provided by users.
However, accurate budget allocation based on channel characteristics is a difficult problem for the top-level strategy, particularly when dealing with complex channel distributions (i.e., the customer traffic varies in different channels at different times).
To this end, a CPC-constrained diffusion budget allocation model is proposed by incorporating an additional CPC-based loss into the optimization objective of  diffusion regularization methods, in order to satisfy the CPC constraint during the budget allocation learning process. 
The CPC-constrained diffusion model offers several advantages: (1) it can capture the multi-modal distribution of budget allocations in dynamic environments, particularly in RTB, where advertiser budgets, market conditions, and customer traffic fluctuate significantly over time; and (2) it can incorporate value function maximization and CPC constraints at each reverse diffusion step to enable more effective budget allocation.

\textbf{At the bottom level}, we first propose a novel offline RL approach, namely the state-action decoupled actor-critic method, which learns a state policy to predict the optimal next state and an action policy to infer the action given the predicted next state.
In this way, we can decompose the state-action learning
in the original task into state learning and action learning, and conduct these two learning processes independently in a supervised manner, thereby avoiding the extrapolation error problem caused by out-of-distribution actions in traditional offline RL.
Moreover, to leverage cross-channel information to improve bidding strategies and mitigate the potential data scarcity problem, we propose a context-based meta-channel knowledge learning method by introducing a channel-shared meta learning objective that extracts the shared knowledge from different channels to learn a more generalizable channel representation. 
Finally, formulating the bidding strategies for different channels as a multi-agent RL problem, we utilize a central value function based on the observations from all channels to achieve global maximization of bidding performance.
The main contributions of this paper are as follows:
\begin{itemize}

    \item We provide a formal RL formulation for the problem of cross-channel bidding with a shared budget, and propose a novel solution HMMCB to optimize cross-channel bidding decisions with dynamic budget allocations from a hierarchical and multi-agent perspective.
   
    \item We propose a state-action decoupled actor-critic method in offline RL and a knowledge-sharing context-based learning method in meta RL, in order to utilize the cross-channel interrelationships for efficient bidding decision making within each channel.
    
    \item  Empirical results from offline evaluation and online A/B  tests  on a large scale realistic ad bidding platform prove the superiority of HMMCB over state-of-the-art methods.

\end{itemize}

\section{Related work}

\subsection{ Bidding Optimization in RTB }

The objective of bidding optimization is to determine the appropriate bid price for each impression presented in the auction to achieve total revenue maximization. 
Perlich et al.~\cite{DBLP:PerlichDHSRP12} first introduce a linear bidding strategy based on ad impression evaluation, which has been widely adopted in real-world applications. 
Zhang et al.~\cite{DBLP:ZhangYW14} and Wang et al.~\cite{DBLP:WangYYZLS19} propose bidding optimization approaches based on the nonlinear relationships between optimal bids and ad impression evaluation. 
In addition, many works model bidding optimization from a sequence decision perspective and solve it using RL methods~\cite{jiang2023adaptive, tunuguntla2023bidding,DBLP:WuCYWTZXG18,DBLP:CaiRZMWYG17}.
Cai et al.~\cite{DBLP:CaiRZMWYG17} utilize an MDP framework to sequentially allocate budget based on real-time impressions.
Wang et al.~\cite{DBLP:WangDFY0WZ22} develop a course-guided Bayesian reinforcement learning (CBRL) framework to adaptively balance the multiple constraints in a non-stationary advertising market.
Tunuguntla et al.~\cite{tunuguntla2023bidding} propose a multi-period dynamic programming model to provide advertisers with the optimal combination of generic and brand bids for search ads.

Most existing studies concentrate on bidding optimization under single-channel setting and cannot be directly applied on scenarios where there are multiple ad channels with significant quality differences.
On this basis, some studies~\cite{DBLP:XiaoGJLCZY19, DBLP:www/AlonGT12, DBLP:JinSLGWZ18, DBLP:WenXZZWLRXTYXWC22, DBLP:WangTLMZDSXWW24 } have developed methods for allocating budgets across multiple advertising channels under total budget constraints via model-base RL or influencer probabilistic modeling.
However, these methods require an accurate distribution estimation of bidding results, such as the expected number of clicks, and overlook the interconnections between different channels, which restricts their applicability. 
Moreover, Jin et al.~\cite{DBLP:JinSLGWZ18} address bidding optimization across various channels through multi-agent reinforcement learning.
However, they adopt a budget allocation scheme that shares the total budget as a whole across different channels, which can lead to a greedy budget allocation and thus globally suboptimal performance.
While wen et al.~\cite{DBLP:WenXZZWLRXTYXWC22} investigate the competition and cooperation relation among auto-bidding agents and propose a temperature-regularized credit assignment to establish a mixed cooperative-competitive paradigm, they did not consider the issue of channel budget allocation.
Wang et al.~\cite{DBLP:WangTLMZDSXWW24} employ $\lambda$-generalization to adapt to budget changes across different channels, treating each channel as an independent agent and aiming to achieve global optimality through individual optimal values.
However, the assumption that channels will not affect each other does not hold in multi-channel advertising bid setting, thus limiting its performance.
Unlike previous works, we utilize each channel's characteristics and interrelationship information across multiple bidding channels to achieve dynamic budget allocation and optimal bidding decisions.

\subsection{RL}

\textbf{Offline RL} allows for the learning of effective policies solely from pre-existing data, thus avoiding potential risks induced by online interactions. 
However, offline RL encounters well-known distribution shift challenges~\cite{levine2020offline}, which is typically attributed to evaluation of actions that do not exist in the offline dataset (i.e., out-of-distribution actions) during policy learning.
To mitigate this problem, recent studies have proposed several policy regularization methods that force the learned policy to stay close to the behavior policy by equipping policy optimization objective with various regularizers, including different divergence penalties~\cite{DBLP:abs-1911-11361, DBLP:KumarFSTL19, DBLP:FujimotoG21}, implicitly weighted behavior cloning~\cite{DBLP:0001NZMSRSSGHF20, DBLP:abs-1910-00177}, or reconstruction loss of the generative policy model~\cite{DBLP:FujimotoHM18, DBLP:WangHZ23}. 
In contrast to directly regularizing the learned policies, an alternative approach is to incorporate behavioral regularization into value estimation objective, encouraging adherence to the behavioral distribution through a conservative value estimate towards unseen state-action pairs~\cite{DBLP:abs-1912-02074, DBLP:XuZZ22}. 

The above methods require the setting of complex regularization terms to ensure the effectiveness of strategy learning, which becomes challenging for cross-channel advertising bidding platforms with diverse distribution changes (i.e., customer traffic, ad format and advertising request time).
Our method is a simpler regularization method that decouples the process of learning behavior policy into state learning and action learning, which adopts an imitation-style manner to train the state policy and action policy.
By doing so, our method can model state-state correlation, rather than state-action correlation in previous work, which avoids the extrapolation errors of evaluating out-of-distribution actions.

\textbf{Meta RL} equips agents with the capability to quickly adapt to novel tasks through training on a diverse task distribution.
Optimization-driven methods~\cite{DBLP:FoersterFARXW18, DBLP:HouthooftCISWHA18} are one type of the most widely-studied approaches to achieve this goal, which formalizes the task adaptation process by employing policy gradients over a limited set of few-shot samples, thereby acquiring an optimal policy initialization.
In addition, context-based Meta RL methods~\cite{DBLP:ZintgrafSISGHW20, DBLP:FakoorCSS20} model the meta-learning problem as a partially observable Markov decision process (POMDP), where the task information is viewed as hidden components of states and extracted from historical trajectories.

The above methods focus on the distribution mismatch between offline data and online exploration data, and require additional online exploration.
Unlike them, we consider using entirely offline data to learn channel common  knowledge representation and leverage this common knowledge to enhance the learning effectiveness of the strategy.

\begin{figure}[t]
    \centering
    \includegraphics[scale=0.65]{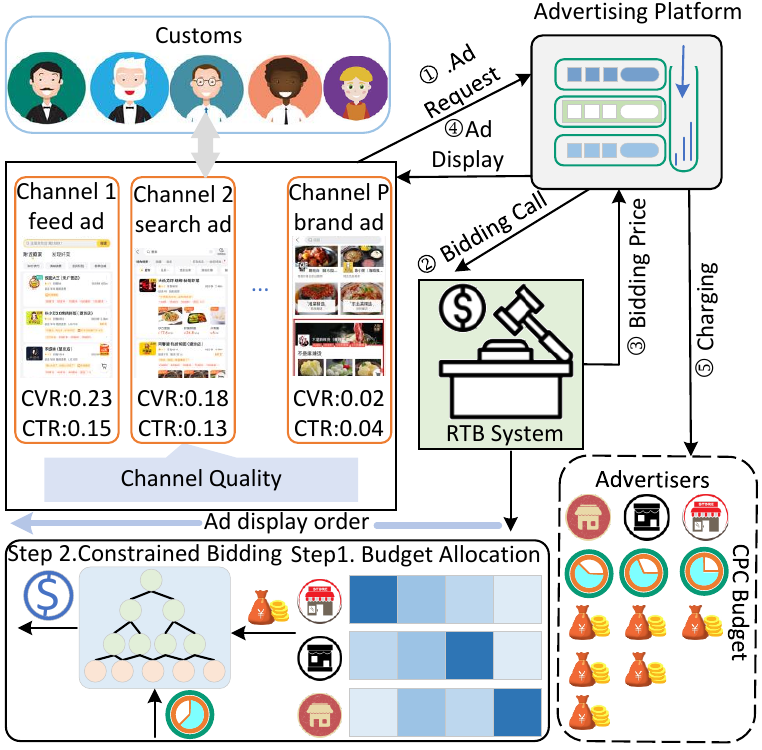}
    \caption{An overview of the Meituan  advertising system. Advertising platforms need to bid on different channels based on the budget and constraints of merchants. CTR and CVR represent average click-through rate and conversion rate respectively.}
    \label{fig:bidding_scen}
\end{figure}

\section{Preliminary}

\subsection{Reinforcement Learning}
A reinforcement learning (RL) problem is a sequential decision-making problem, which is generally modeled as a Markov Decision Process (MDP) represented by a tuple $(S,A, \mathcal{P},R)$, where $S$ is the state space of the environment with each state $s \in S$ representing a unique situation within the environment, 
$A$ denotes the set of all actions that an agent can take within the environment, 
$R : S \times A \rightarrow \mathbb{R}$ is the numerical reward obtained as a result of a particular action taken within a specific state, and $\mathcal{P}: S\times A \rightarrow [0, 1]$ is the probabilistic transition function that captures the impact of an action on the future state, where $s_{t+1} \in S$ is the next state that is observed after taking action $a_t \in A$ in a given state $s_t \in S$ at a time step $t$. 
The agent takes actions according to a policy $\pi : S \times A \rightarrow [0, 1]$, which maps the states to a probability distribution over actions.
The goal of RL is to learn a policy $\pi$ that maximizes the expected total reward in the environment: 
\begin{equation}
    \max_{\pi}J(\pi) =\max_{\pi} \mathbb{E}_{\mathcal{P}, \pi}\left[\sum_t R(s_t,a_t)\right]
\end{equation}

In the offline RL setting where online interactions are inaccessible~\cite{DBLP:abs-1911-11361, DBLP:abs-1912-02074, DBLP:FujimotoG21}, the goal is to maximize the expected total reward using only the offline dataset $D=\{(s_t,a_t,s_{t+1},r_t)\}^T_{t=1}$ collected by a behavior policy.
Training a policy that outperforms the behavior policy using historical data often entails querying the value function of the actions that were not seen in the dataset (i.e., out-of-distribution actions).
These actions can be viewed as adversarial examples for the Q function~\cite{DBLP:KumarZTL20}, leading to an extrapolation error of value estimation.

\begin{figure*}[htbp]
    \centering
    \includegraphics[scale=0.73]{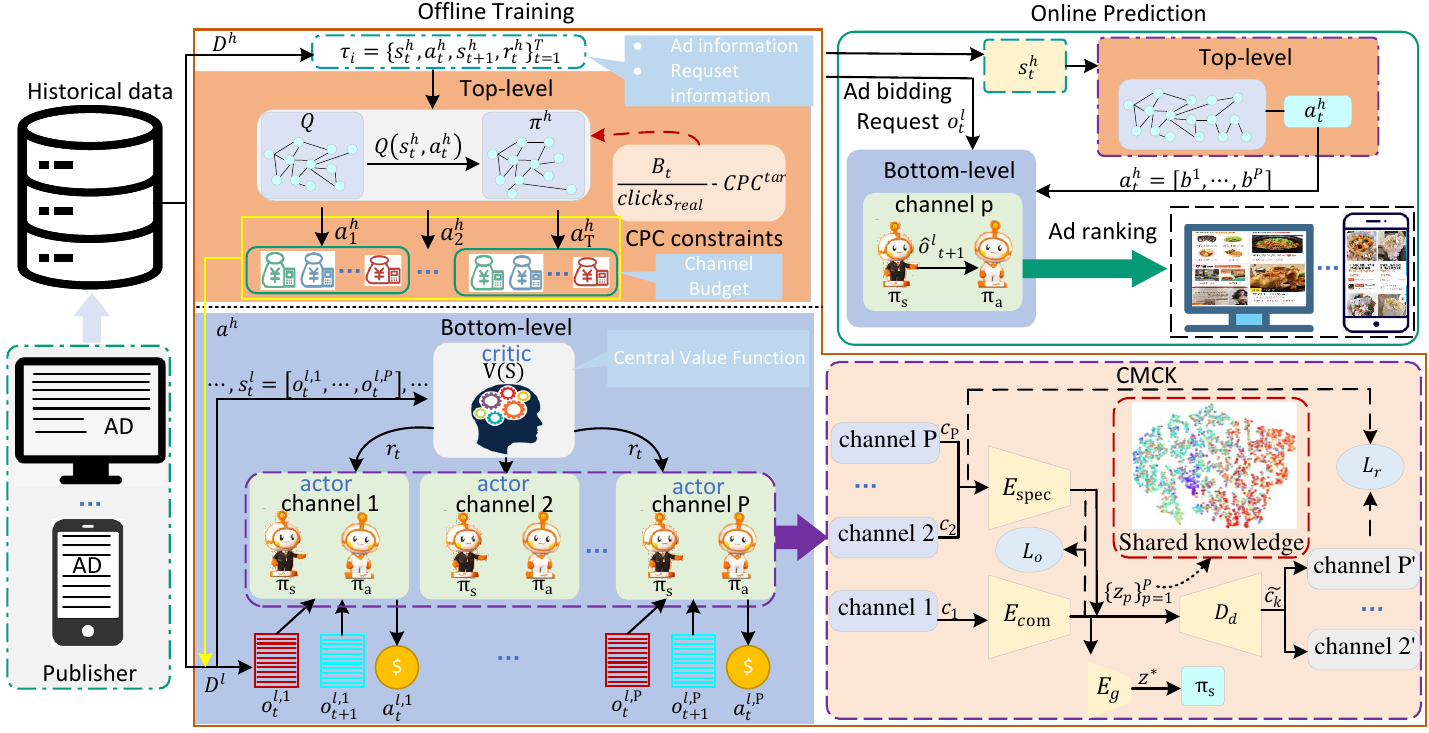}
    \caption{The HMMCB framework.}
    \label{fig:framework_img}
\end{figure*}

\subsection{Meituan Cross-channel Budget Sharing Ad System}

Boasting over 400 million users and 66 million merchants, Meituan\footnote{https://www.meituan.com/en-US/about-us} is one of China’s largest ad bidding platforms for online services including food delivery, hotel and travel et al., which handles an average of more than 100 million impression requests every day.
The overall process of Meituan advertising system is shown in Fig.~\ref{fig:bidding_scen}.
When receiving an advertising impression request, the advertising platform sends a bidding call to the RTB system.
The RTB system then supports advertisers in determining a bidding price for this impression based on the real-time information of each channel as well as the budgets and bidding constraints specified by advertisers.
After that, the advertising platform shows ads from the winning bidding groups and deducts the respective bid amounts from their budgets (i.e., the displayed advertiser will be charged the price of the second highest bid in this ad auction).
Finally, this advertiser will receive revenue if a consumer clicks on the ad and makes a purchase.

A significant feature of the Meituan ad system is that advertisers can place ads on one or more channels with a shared total budget that is updated every day.
Because ad requests arrive at different times for each channel, the channel that receives the earliest ad request will send the ad bid.
With a sufficient budget, advertisers often bid above the actual bidding price to get the ad slot, particularly for recommended and brand ads that do not require user activation.
However, this may result in advertisers not having enough funds to invest in other channels (e.g., the search ad channel) later in the day, causing them to miss out potential users and ultimately reducing the revenue.
While allocating a budget for each individual channel can address this issue, the complexity arises from varying budget requirements based on channel characteristics and the need for different budget amounts at different times due to the fluctuating customer traffic.

In addition, making optimal bidding decisions in cross-channel bidding requires considering information from multiple channels.
For example, consider two advertisers, $A$ and $B$, bidding on two channels, $C_1$ and $C_2$. Assume that $A$ can successfully bid on both channels, while $B$ can only succeed on $C_2$ when $A$ is not bidding, and the advertising revenue for $A$ from $C_1$ significantly outweighs that from $C_2$. 
In a scenario where $C_1$ and $C_2$ initiate ad bidding simultaneously, if $A$ wins bids on both channels, $B$ loses the opportunity to display ads on $C_2$, resulting in decreased revenue. 
Conversely, if $A$ bids only on $C_1$ and $B$ on $C_2$, both $A$ and $B$ can obtain corresponding revenues. 
Thus, effective bidding strategies should consider not only the characteristics of each channel but also the combined bidding information of multiple channels in order to achieve optimal decisions.

\subsection{Optimization Objective}
In this paper, our overall objective is to maximize total ad clicks while satisfying all advertisers' set budgets and CPC constraints, ensuring that the platform's revenue remains within an acceptable range.
As shown in Fig.~\ref{fig:framework_img}, our approach is divided into a top-level strategy and a bottom-level strategy.
The goal of top-level strategy is to allocate budget to each channel while meeting the advertisers's total budget.
Consider a scenario where $M$ advertisers are served by a top-level strategy that allocates the budget of each advertiser to $P$ channels. Each advertiser $m$ provides a budget $\mathcal{B}_m$ to the advertising platform once a day, and then the top-level strategy allocates this budget to different channels to constrain the bidding decisions therein. 
Formally, the objective of the top-level strategy can be expressed as follows:
\begin{equation}
    \begin{split}
       & \mathop{\max}_{b^{p}_{m,t}}\sum^M_{m=1} \sum^P_{P=1} \sum^\mathbb{T}_{j=1} \text{click}(b^{p}_{m,j}) \\ 
       & \text{subject \quad to} : \sum^P_{p=1} \sum^\mathbb{T}_{j=1} b^{p}_{m,j} \leq \mathcal{B}_m, \forall{m} \in \{1,2,\cdots,M\} \\ 
       & \text{and} \quad CPC^{real}_m \leqslant CPC^{tar}_m,
    \end{split}
\end{equation}
where $b^{p}_{m,j}$ denotes the allocated budget on channel $p$ for advertiser $m$ at time $j$, $\text{click}(b^{p}_{m,j})$ represents the number of clicks given the budget $b^{p}_{m,j}$ at time $j$, while $CPC^{real}$ and $CPC^{tar}_m$ represent the actual CPC realized by the advertiser $m$ and the CPC target set by the advertiser $m$, respectively.
To avoid the abuse of symbols, we use $b_j$ to represent the budget of an advertiser at time $j$ across all channels.
Table~\ref{tab:notion} shows a more detailed description of the symbols.

After receiving the budget allocation $b$ from the top-level strategy, the bottom-level strategy  then maximizes the number of clicks while satisfying the budget. 
Assume there are $T$ ad requests on channel $p$ during the time interval, the objective of the bottom-level strategy can be expressed as follows:
\begin{equation}
    \label{eq:low_return}
    \begin{split}
        & \mathop{max}_{a^{p}_{t}} \sum^P_{p=1} \sum^{T}_{t=1} click(a^{p}_{t}) \\
        & subject \quad to: \sum^{T}_{t=1} cost(a^{p}_{t}) \leq b^{p}, \quad \forall p \in \{1,\cdots,P\}, \\
    \end{split}
\end{equation}
where $cost(a^{p}_{t})$ denotes the actual cost, $b^p \in b$ represents the budget of channel $p$, and $click(a^{p}_{t})$ indicates whether the user has clicked the ad after giving a bidding price $a^{p}_{t}$. 
Note that, at the bottom-level, we take the number of ad requests as $\mathbb{T}$. Hence, $t$ represents both the number of ad requests and the time instance for the bottom-level strategy.

\section{Method}
The overall framework of HMMCB is shown in Fig.~\ref{fig:framework_img}.
During  \textbf{the offline training process}, the top-level strategy applies the \textit{CPC-constrained diffusion budget allocation method} to achieve dynamic budget allocation among the channels while satisfying the CPC constraint, and the bottom-level strategy utilizes the \textit{state-action decoupled actor-critic method}, \textit{context-based meta-channel knowledge learning method} and \textit{multi-agent RL training scheme} for offline RL, cross-channel knowledge sharing and cross-channel bidding decision making, respectively.

During  \textbf{the online prediction process}, the top-level strategy allocates the total budgets to each channel based on the global information from all channels. 
Meanwhile, the bottom-level strategy for a specific channel makes bidding decisions based on the allocated budgets and real-time channel information.
Finally, the ads of the successful bidders will be displayed in the ad space of that channel. 

\subsection{Top-level Strategy for Dynamic Budget Allocation}

\begin{table}[]
\centering
\caption{Important symbols and explanations in this paper}
\begin{tabular}{ll}
\hline
Notation       & Explanation                                                                                                              \\ \hline
M,m            & total number, index of advertisers                                                                                       \\ \hline
P,p            & total number, index of channel                                                                                           \\ \hline
$\mathbb{T}$,j            & total number, index of time step                                                                                         \\ \hline
T,t            & total number, index of ad request                                                                                        \\ \hline
ROI, CPC       & return-on-investment, cost-per-click                                                                                     \\ \hline
$D^h$,$D^l$          & top-level and bottom-level datasets                                                                                      \\ \hline
$s$,$\hat{s}$          & \begin{tabular}[c]{@{}l@{}}top-level and bottom-level \\ global observation (state)\end{tabular}                         \\ \hline
$\mathcal{O}$,$b$,$g$       & \begin{tabular}[c]{@{}l@{}}local observation (state), action, \\ reward of time $j$ in top-level MDP\end{tabular}          \\ \hline
$O$,$a$,$r$       & \begin{tabular}[c]{@{}l@{}}local observation (state), action \\ and reward of ad request t in bottom-level MDP\end{tabular} \\ \hline
$B$            & total budget                                                                                         \\ \hline
$Q_\phi$, $\pi_\theta$     & top-level Q network and policy network                                                                                   \\ \hline
$V_\phi$,$\pi_s$,$\pi_a$ & \begin{tabular}[c]{@{}l@{}}bottom-level V network, state and \\ action policy network\end{tabular}                       \\ \hline
\end{tabular}
\label{tab:notion}
\end{table}

\textbf{1) MDP formulation for budget allocation}.
The top-level MDP for each advertiser $m$ is formulated as a tuple $(\mathbb{S}, \mathbb{B}, \mathbb{P}, \mathbb{R}, \gamma)$, where $\mathbb{S}$ is the state space, $\mathbb{B}$ denotes the action space, $\mathbb{P}$ indicates the state transition function, $\mathbb{R}$ represents the reward function, and $\gamma$ is the discount factor. 

\begin{itemize}
    \item \textbf{State $\mathbb{S}$}: the $s \in \mathbb{S}=\{\mathcal{O}^{1},\cdots,\mathcal{O}^{P}\}$ contains the state of each channel, where $\mathcal{O}^{p}$ denotes the state space of the \textit{p}-th channel, which comprises the advertiser allocated budget and historical statistics (e.g., the user-preferences, average click-through rate (CTR) and conversion rate (CVR) ). 
    Note that an advertiser might engage in bidding activities exclusively in specific channels, with some channels lacking information related to this advertiser. 
    To maintain data consistency and training stability, we employ zero tensors of identical dimensionality to supplement $s$.
    \item \textbf{Action $\mathbb{B}$}: the $ b \in \mathbb{B}$ encompasses the budget for each individual channel, where $b^p$ represents the budget allocated to channel $p$. 
    We discretize the action using the percentage of the budget and mask invalid actions that exceed the total budget. 
    \item \textbf{Reward $\mathbb{R}$}: the reward $g \in \mathbb{R}$ is calculated as the sum of the number of user clicks across all channels within a given time interval.
\end{itemize}

According to the above formulation, we obtain offline data $D^h=[\tau_1,\tau_2,\cdots,\tau_N]$ from the bidding log, where $\tau_i=\{s_j,b_j,g_j\}^T_{j=1}$ \footnote{We set $T=7$ in compliance with the Meituan system, corresponding to a week of advertising bidding activities} denotes an offline trajectory.

\textbf{2) CPC-constrained diffusion model for dynamic budget allocation.}
Multi-channel advertising bidding is a dynamic environment with tens of thousands of advertisers and customers participating at every moment.
In this scenario, the policy must be expressive and accurately capture the multimodal distribution~\cite{DBLP:WangHZ23} to dynamically allocate an appropriate budget for each channel.
To achieve this, we employ a policy regularization method that utilizes a diffusion model in the action space, forming a conditional diffusion model conditioned on states~\cite{DBLP:SongE19, DBLP:HoJA20, DBLP:WangHZ23}. 
In specific, the top-level strategy is expressed through the inverse process of a conditional diffusion model, denoted as: 
\begin{equation}
\label{eq:top_poliy}
    \begin{split}
        \pi_\theta(b|s) &=p_\theta (b^{0:\mathcal{I}}|s) \\
        &= \mathcal{N}(b^{\mathcal{I}};0,I) \Pi^\mathcal{I}_{i=1}p_\theta(b^{i-1}|b^{i},s),
    \end{split}
\end{equation}
where the end sample of the reverse chain (i.e., $b=b^{0}$) is the action used for RL, $i \in \{0,\cdots,\mathcal{I}\}$ denotes the diffusion timestep, and
$\mathcal{N}$  represents Gaussian distribution. 
To train the conditional model, we follow the simplified objective set by DDPM~\cite{DBLP:HoJA20} and Diffusion-QL~\cite{DBLP:WangHZ23} as follows:

\begin{equation}
    \begin{split}
    \label{eq:lead_loss}
         L_{simple}(\theta):=\mathbb{E}_{i \sim \mu,\epsilon \sim \mathcal{N}(0,1),(s,b) \sim D}[||\epsilon-\\ \epsilon_\theta(\sqrt{\bar{\alpha_t}}b+\sqrt{1-\Bar{\alpha_i}}\epsilon,s,i)||^2].
    \end{split}
\end{equation}
Thus, 
\begin{equation}
\label{eq:df_loss}
    \begin{split}
        \pi^h &=\mathop{argmax}_{\pi_\theta} \mathcal{L}(\theta) \\
        &=L_{simple}-\alpha \cdot \mathbb{E}_{s \sim D^h, b^{0} \sim \pi_\theta}[Q_\phi(s,b^{0})],
    \end{split}
\end{equation}
where $\epsilon$ is a random variable, $\alpha$ is a hyperparameter, and $Q$ is the $Q$-value network for policy evaluation. 

While maximizing returns (clicks) may expand the potential customer groups, there is a risk of running out the advertising budget rapidly if the \text{CPC} becomes excessively high.
By setting a \text{CPC} constraint, advertisers can better control their advertising budget, ensuring the sustainability of their advertising bidding.
However, although the use of hard constraints~\cite{garcia2015comprehensive} can force the policy to stay within the \text{CPC} constraints set by the advertiser, it may also encounter the problem that no feasible policies can be found to meet the constraints, thus impacting the final learning efficiency and performance.
To avoid this challenge, we transform the \text{CPC} constraint into a variance loss and incorporate a reverse diffusion chain during the training phase.
Assuming that all bottom-level strategies are fully capable of conducting bidding activities in accordance with the budget allocated by the top-level strategy, the actual $CPC^{real}$ can be expressed as:
\begin{equation}
    CPC^{real}=\frac{\sum_{p=0}^P(b^p)}{g}.
\end{equation}
If we consider $CPC^{real}$ as the output label and $CPC^{tar}$ as the target label, the \textit{CPC} constraint problem can be simplified into a linear regression problem and thus be optimized using the 
mean square errors of $CPC^{tar}$ and $CPC^{real}$ as:
\begin{equation}
\label{eq:cpc_loss}
    L_{\text{CPC}}=\frac{\mathop{\sum}_{m=1}^M (CPC^{tar}-CPC^{real})^2}{M}.
\end{equation}
Equation~(\ref{eq:cpc_loss}) can be approximated as a soft constraint~\cite{huang2023posterior}, which ensures that the diffusion policy can maximize the exploration around the $CPC^{tar}$ by minimizing the variance of the advertiser's CPC constraint.
Finally, the ultimate policy-learning objective comprises a linear combination of policy regularization, CPC constraint and policy enhancement components, as follows:

\begin{equation}
\label{eq:top_loss}
    \begin{split}
        \pi^h &=\mathop{argmax}_{\pi_\theta} \mathcal{L}(\theta) \\
        &=L_{simple}+L_{\text{CPC}}-\alpha \cdot \mathbb{E}_{s \sim D^h, b^{0} \sim \pi_\theta}[Q_\phi(s,b^{0})].
    \end{split}
\end{equation}
The procedure for offline training of the top-level algorithm is delineated in Appendix $ \text{\uppercase\expandafter{\romannumeral2}}$.

\subsection{Bottom-level Strategy for Cross-channel Bidding}
\textbf{1) Bottom-level MDP for cross-channel constrained bidding}.
Since each request $t$ only affects the cost of a specific channel $p$, we model each channel separately and represent it as a tuple$(S,A,P,R,\gamma)$.

\begin{itemize}
    \item \textbf{State $O$}: the $o \in O$ state contains the allocated budgets $b^h$, bidding requests, and advertiser information, where the bidding request comprises the request time and current advertising status (e.g., the budget consumption rate and the ratio of financial constraint satisfaction), while the advertiser information is identical to the state of the top-level strategy ($\mathcal{O}$).
    \item \textbf{Action $A$} : the $a \in A$ is a bidding ratio, and the final bidding price is calculated using $a_t \times CPC^{tar}_m$.  
    Note that $ a \in [\xi_{min}, \xi_{max}]$, while $\xi_{min}$ and $\xi_{max}$ have various values in different channels.
    \item \textbf{Reward $R$}: the $r \in R$ represents the number of user clicks on the ad. For each request $t$, if the bidding is successful and the user finally clicks the ad, the reward $r_t \in \{0,1\}$ is set to 1. 
    However, because using clicks as a reward may lead to costs exceeding a budget of the channel, we introduce a budget constraint
    \begin{equation}
        c_t=cost(a_t)-b^{p},
    \end{equation}
    which ensures to maximize ad clicks while keeping the costs within predefined budget limits. 
    Finally, the bottom-level reward is defined as follows:
    \begin{equation}
            r_t=click(a_t)-c_t.
    \end{equation}
\end{itemize}

According to the above formulation, we obtain offline data $D^l=[\tau_1,\tau_2,\cdots,\tau_N]$ from the bidding log, where $\tau_i=\{o_t,\cdots,o^{P}_t,a_t,\cdots,a^{P}_t,r_t,\cdots, r^{P}_t\}^T_{t=1}$ \footnote{$T$ represents the total number of ad requests from the start to the end of a day’s bidding, which varies from day to day} denotes an offline trajectory for bottom level strategy learning. 

\textbf{2) State-action decoupled actor-critic for bidding strategy learning}.
In advertising bidding systems where users can freely participate and exit, existing works face challenges in addressing the problem of out-of-distribution actions because the data distribution of bidding logs cannot cover the true bidding data distribution.
We propose the state-action decoupled actor-critic method to overcome this problem by training a state value function, denoted as $V(o): O \rightarrow \mathbb{R}$, that only uses the samples without information of actions, i.e., $(o_t, o_{t+1})$, to evaluate and predict the optimal next state, and an action policy, denoted as $\pi_a(o_t, o_{t+1}): O \times O \rightarrow [\xi_{min}, \xi_{max}]$, to infer the action given the predicted next state.

In order to approximate the optimal value function, we use the asymmetric least squares method from recent research~\cite{DBLP:KostrikovNL22, DBLP:MaYH0ZZLL22}, which involves applying an $ l_2 $ loss with a different weight using expectile regression, resulting in the following asymmetric $ l_2 $ loss function:

\begin{equation}
    \label{eq:low_obs_V}
    \begin{split}
        \mathop{max}_\phi \mathbb{E}_{(o_t,r,o_{t+1}) \sim D}[L^\varrho_2 (r+\gamma V' (o_{t+1})-V (o_t)],
        \\ where \quad L^{\varrho}_2(\mu)= | \varrho-\mathds{1}(\mu \leq 0) | \mu^2.
    \end{split} 
\end{equation}
When $\varrho=0.5$, the operator simplifies to the Bellman expectation operator, whereas when $\varrho=1$, the operator increasingly resembles the Bellman optimality operator. 
Note that our learning objectives are similar to IQL~\cite{DBLP:KostrikovNL22}, but our goal is to learn the state value function, which avoids evaluating state$-$action pairs that are not present in the dataset (i.e., extrapolation errors).

Simply maximizing the state policy $\pi_s$ with respect to $V$ can lead to the state policy overlooking the logical coherence between pairs of states (i.e., it is not possible to reach the next state from the current state), which can hinder the action policy $\pi_a$ from making correct decisions.
To mitigate this potential issue,  a behavior cloning term is incorporated into the learning objective of the $\pi_s$ as follows:
\begin{equation}
\label{eq:g_loss}
    \mathop{max}_{\omega} \mathbb{E}_{(o_t,o_{t+1} \sim D^l)} [log \pi_{s}(o_{t+1} | o_t) +\lambda 
 \cdot V(\pi_{s}(o_t))],
\end{equation}
where $\lambda$ is a hyperparameter for balancing behavior cloning and value guidance, while $\omega$ is the network parameter of $\pi_s$.

During the training phase, the action policy $\pi_a$ performs supervised learning by maximizing the likelihood of the actions given the states and next states, yielding the following objective:
\begin{equation}
\label{eq:e_loss}
    \mathop{\max}_{\theta}\mathbb{E}_{(o_t,a_t,o_{t+1}) \sim D^l}[\log \pi_{a}(a_t|o_t,o_{t+1})],
\end{equation}
where $\theta$ represents the network parameter of $\pi_a$.
During the evaluation phase, the state policy $\pi_s$ generates an optimal next state based on the current state, and 
$\pi_a$ takes action based on the current and next state as:
\begin{equation}
    \label{eq:low_policy}
    a_t=\mathop{\arg\max}_{a_t} \pi_a(a_t|o_t,\pi_s(o_t)).
\end{equation}

\textbf{3) Context-based meta-channel knowledge learning for cross-channel bidding}.
Within the framework of the multi-channel advertising bidding system, each channel exhibits unique characteristics such as different customer types, various advertisement formats (e.g., video and text ads), and distinct ROI metrics. Provided there is sufficient training data, we can learn effective bidding strategies directly based on the information of each channel.
However, the historical data distribution of channels with sparse customer traffic could greatly differ from the real data distribution (i.e., the data scarcity problem), which causes the policy to converge towards the local rather than the global optimal. To solve this problem, we propose a context-based meta-channel knowledge learning method (the \textbf{CMCK} component shown in the right corner of Fig.~\ref{fig:framework_img}), which uses the context-based meta-RL to learn generalizable representations of different channels while transferring common knowledge across channels to facilitate the learning efficiency in each channel.

Specifically, CMCK comprises a \textbf{task-infer module} and a \textbf{policy learning module} that are trained in two phases.
First, the task inference module is trained to encode shared knowledge across channels using two distinct encoders, namely common encoder $E_{com}$ and specific encoder $E_{spec}$. 
The encoder $E_{com}$ is used to encode the information that is shared across the channels, while $E_{spec}$ is employed to encode the information of the task within the specific channel. 
Subsequently, the data are reconstructed by a decoder $D_{d}$ to ensure the integrity of the shared knowledge. 
Then, in the policy learning module, the pre-trained encoder $E_{spec}$ is employed to extract the latent features of the data, and an aggregator $E_g$ is used to aggregate all features to facilitate the learning of the state policy.

Assuming that the bidding data of $p$ channels follow the distribution $\rho(\mathcal{T})$, we sample the batch context $c_p$ (i.e., $c_p$ represents bidding data at different times) of the current channel data $\mathcal{T}_p$ and another context $c_k$ sampled from the other $k$ channels $\mathcal{T}_k \in \mathcal{T} \setminus \{\mathcal{T}_p\}$. 
The task-infer module is comprised of $\tilde{c}_k=D(E_{com}(c_p)+E_{spec}(c_k))$ and $z=E_{com}(c_p)$, where $\tilde{c}_k$ represents the reconstruction of inputs $c_p$ and $c_k$, while $z$ symbolizes the shared knowledge. 
Let $z_p$ and $z_k$ denote the matrices of the features of $c_p$ and $c_k$, respectively. 
The soft subspace orthogonality loss is used to facilitate the formation of distinct representations by the common and specific encoders within the latent space $z$ as follows:
\begin{equation}
    L_o=\|z_p^T z_k\|^2_F,
\end{equation}
where $\|*\|^2_F$ represents the squared Frobenius norm. 
In addition, the reconstruction loss $L_r$ is used to guarantee that the shared knowledge $z_p$ is transferred between different channels:
\begin{equation}
    L_r=\mathop{\sum}_{p=0}^{P} L_{smse}(c_k,\tilde{c}_k),
\end{equation}

\begin{equation}
\label{eq:meta_loss}
    L_{smse}(c,\tilde{c})=\frac{1}{n}\|c - \tilde{c}\|^2_2 -\frac{1}{n^2}((c-\tilde{c})\cdot 1_n)^2,
\end{equation}
where $\|*\|^2_2$ denotes the squared $L_2$-norm, $n$ is the number of element in input $c$, and $1_n$ is the vector of ones of length $n$. 
The objective of the CMCK training process is to minimize the following loss concerning the parameters $\Theta=\{\theta_{com},\theta_{spec},\theta_d\}$:
\begin{equation}
    L=L_o + \eta \cdot L_r,
\end{equation}
where $\eta$ is a weight hyperparameter.

When the state policy $\pi_{s}$ is updated, CMCK expands the state $s$ by potentially representing $z^*$, which improves the adaptability of policy $\pi_{s}$ to effectively respond to changes across different channels.

\textbf{4) Multi-agent training for  cross-channel bidding.}
Training each channel strategy independently without considering their interconnections can lead to suboptimal results.
To this end, we model the bottom-level strategy as a cooperative multi-agent to alleviate the local optimal problem of independent training channel strategy.
Specifically, the bottom channel bidding can be viewed as a Partially Observable Markov Decision Process (POMDP)~\cite{DBLP:abs-1301-3836, DBLP:ChenLJ24} $\mathcal{G}= ⟨N,\hat{S},\hat{A},\{R_i\}_{i \in P}, \mathcal{P}⟩$, where $P$ represents the number of channels, $\hat{S}=[o^1,\cdots,o^P]$ denotes the joint state space, $\hat{A}=[a^1,\cdots,a^P]$ indicates the joint action space, $\mathcal{P}:S \times A \rightarrow S$ represents the transition function, and ${R_i}$ is the reward functions for channel strategy $i$. 
Our learning objective is:
\begin{equation}
    \mathop{\text{max}}_{\pi_{ent}} \mathbb{E} [\sum_{t=0}^{T} \gamma^t(R(o_t,a_t)-\mathcal{K} f(\pi_{ent}(a_t|o_t),\pi^*_{ent}(a_t|o_t)))],
\end{equation}
where $\mathcal{K}$ is a scale parameter, $\pi_{ent}$ is the joint learning policy, $\pi^*_{ent}$ denotes joint behavior policy, and $f(\cdot)$ represents the function that captures the divergence between $\pi_{ent}$ and $\pi^*_{ent}$.
However, the state-action space of multi-channel scenario is enormous, making it difficult to directly compute the regularization term between the global policy \( \pi_{\text{tot}} \) and the global behavior policy \( \pi^*_{\text{tot}} \).
Although existing works~\cite{DBLP:YangMLZZHYZ21, DBLP:PanH0X22, DBLP:WangXZZ23} use local regularization instead of global regularization by decomposing the function $f$, local regularization cannot be effectively applied to other channels due to the huge gap in state features (e.g., user traffic and ad types) between each channel.

To this end, we use central value functions to guide the learning of each strategy (i.e., global value functions guide local regularization), which aligns with the centralized training with decentralized execution (CTDE) framework~\cite{DBLP:LoweWTHAM17}.
Simply put, each channel is considered an agent, and the training process requires all $P$ learning agents to participate in jointly improving the policy network parameters $\theta$ and value network parameters $\phi$. 
At time $t$ and $t+1$, the value function gathers observations from all agents, denoted as $\hat{s}_t=[o^1_t,\cdots,o^P_t]$ and $\hat{s}_{t+1}=[o^1_{t+1},\cdots,o^P_{t+1}]$, respectively. 
The reward $r_t$ is the sum of rewards from all channels, represented as $\hat{r}_t = r^1_t + \cdots +r^P_t$. 
Thus, Eq.~(\ref{eq:low_obs_V}) can be rewritten as:
\begin{equation}
\label{eq:muti-v}
\begin{split}
  &\mathop{max}_\phi \mathbb{E}_{(s_t,\hat{r},s_{t+1}) \sim D^l}[L^\varrho_2 (\hat{r}_t+\gamma \sum_{p=0}^P V' (o^{p}_{t+1}))-\sum_{p=0}^P V(o_t)] \\
  &=\mathop{max}_\phi \mathbb{E}_{(s_t,\hat{r},s_{t+1}) \sim D^l}[L^\varrho_2 (\hat{r}_t+\gamma  V' (\hat{s}^{p}_{t+1}))-V (\hat{s}_t)].
\end{split}
\end{equation} 
In the absence of ad requests at time $t$ for channel $p$, the state remains unchanged (i.e., $o_t=o_{t-1}$), and the reward is set to zero (i.e., $r^p_t=0$).
Unlike value functions, the bidding policy (i.e., state policy and action policy ) can only make actions based on local observations $o_t$ specific to their channel (e.g., customer traffic, bidding, and the budget for that channel), which can be denoted as:
\begin{equation}
    \mathop{max}_{\omega} \mathbb{E}_{(o^p_t,o^p_{t+1} \sim D^l)} [\sum_{p=0}^P \log \pi_{s,p}(o^p_{t+1} | o^p_t) +\lambda 
 \cdot V(\pi_{s}(\hat{s}_t))],
\end{equation}
\begin{equation}
    \mathop{\max}_{\theta}\mathbb{E}_{(o^p_t,a^p_t,o^p_{t+1}) \sim D^l}[\sum_{p=0}^P \log \pi_{a,p}(a^p_t|o^p_t,o^p_{t+1})].
\end{equation}
The offline training process of the bottom-level strategy is shown in Appendix $ \text{\uppercase\expandafter{\romannumeral2}}$.

Finally, after offline model evaluation, HMMCB can be deployed to the online system for online predictions.
For each ad request $t$, HMMCB obtains advertiser-level features and request-level features (i.e., $s^l$) from the ad platform. Then, the top-level strategy takes advertiser-level information and user-preferences (i.e, $s^h$) as input to allocate budget $b=[b^1,\cdots,b^P]$.
After getting the budget of each channel, the bottom-level strategy gives the final bidding actions according to Eq.~\ref{eq:low_policy}.
To ensure model optimality, HMMCB continuously optimizes and updates the model based on online data and user feedback, with the implementation details outlined in Experiment~\ref{sec:online testing}.

\section{EXPERIMENT}
In this section, we conduct a series of offline and online experiments to assess the performance of  HMMCB. 
We first describe the primary setup, and then compare HMMCB with representative learning models to validate its effectiveness. 
Finally, ablation studies are conducted to verify the efficacy of each key component and hyperparameter in HMMCB.

\subsection{Experiment Setup}
\subsubsection{Dataset} We use log data from the Meituan real-time advertising system for offline training and performance evaluation. 
These data are a mixture of expert, medium and random data, where expert data consists of bids achieving an advertising ROI exceeding the expected ROI while satisfying CPC constraints through an automated bidding strategy, medium data includes bids that either meet CPC constraints or surpass the expected ROI, and random data is bids generated through exploration with a random strategy.
Specifically, these data span 35 days of bidding logs collected from four channels: feed ad, search ad, brand ad, and recommendation ad, with an average sampling of 80 million ad requests from 72,351 advertisers across four channels each day and Table~\ref{tab:log_feat} describing the important features in the logs.
The dataset is divided into two segments for training and evaluation: 28 days of bidding data for training and 7 days for evaluation.

\begin{table}[]
\caption{Features description}
\centering
\begin{tabular}{ll}
\hline
Notation                & Explaination                  \\ \hline
ctr                     & Click-through-rate            \\
histctr                 & Historical CRT                \\
final\_charge           & Final transaction price       \\
cpc                     & Cost-per-click                \\
aimcpc                  & Target cost-per-click         \\
impr                    & Ad impression                 \\
revenue                 & Ad revenue                    \\
clicks                  & User clicks on advertisements \\
order\_num              & Orders number                 \\
mt\_butie               & Meituan subsidies             \\
budget                  & Advertiser's Budget           \\
user\_id                & user ID                       \\
request\_time           & Ad bidding request time       \\
dt                      & Log creation time             \\
pvid                    & Channel ID                    \\
impr\_before1week       & Impression from a week ago    \\
click\_before1week      & User clicks from a week ago   \\
cpc\_before1week        & CPC from a week ago           \\
order\_num\_before1week & User orders from a week ago   \\
...                     & ...                           \\
order\_num\_before4week & ...                           \\
user\_pref              & User preferences              \\ \hline
\end{tabular}
\label{tab:log_feat}
\end{table}

\subsubsection{Simulated offline evaluation system} Implementing a predictive model into an operational system without comprehensively evaluating its potential implications has inherent risks (i.e., financial losses).
To mitigate this issue, we have developed an offline evaluation system for simulating the realistic cross-channel bidding process. 
This system comprises two modules: an advertising system simulator and a user feedback predictor. 
The former simulates the Meituan online advertising platform, including retrieval, bidding, ranking, and pricing, while the latter predicts the user feedbacks on ads and evaluates the results.

\subsubsection{Compared methods}
We compare the recent methods that can or can be adapted to handle multi-channel constraints, categorized into three lines of works: (1) single-channel slotwise approximation methods PID~\cite{DBLP:AngCL05} and CEM~\cite{DBLP:BoerKMR05}; (2) soft combination (RL-based) methods MCQ~\cite{DBLP:LyuMLL22}, CBRL~\cite{DBLP:WangDFY0WZ22} and HiBid~\cite{DBLP:WangTLMZDSXWW24}; and (3) a  hierarchical offline method OPAL~\cite{DBLP:AjayKALN21}. 
\begin{itemize}
    \item \textbf{PID} is a classical feedback controller, known for its effective performance in unknown environments. We utilize it to maintain the advertiser's current CPC close to the target CPC, ensuring compliance with the advertiser's cross-channel CPC constraint.
    \item \textbf{CEM} is a gradient-free stochastic optimization technique widely employed in the industry, which strives to optimize a greedy sub-problem within each time slot, striking a balance between exploration and exploitation.
    We formulate the multi-channel bidding problem as an optimization challenge to maximize the click count within the specified total budget and CPC constraint.
    \item \textbf{MCQ} effectively mitigates the value overestimation effect that arises in out-of-distribution actions by proactively training and adjusting their Q values, which is currently considered a state-of-the-art method in the field of offline DRL.
    We use MCQ to model each channel and set clicks as rewards, thus encouraging the strategy to make bidding decisions that maximize the click count.
    \item \textbf{CBRL} is a recognized state-of-the-art approach in the context of ROI-restricted single-channel bidding, which combines Bayesian techniques with an indicator-augmented reward function designed to dynamically manage the trade-off between constraints and objectives.
    To ensure a fair comparison, we adapt CBRL to the multi-channel ad bidding setting by substituting the ROI constraint with a CPC constraint while keeping its original training process.
    \item \textbf{OPAL} is a hierarchical offline algorithm, where the top-level agent is trained using unsupervised learning to provide a temporal abstraction for the bottom-level agent to improve the final policy optimization.
    OPAL and HMMCB are configured with identical hierarchical settings, i.e., the top-level agent allocates the budget, while the bottom-level agent makes advertising bidding decisions.
    \item \textbf{HiBid} is a novel algorithm for bid budget allocation that incorporates an auxiliary loss function to mitigate the risk of over-allocation to specific channels. The algorithm employs $\lambda$-parametric generalization to adapt to variations in budgetary constraints, and integrates a CPC guided action selection mechanism to satisfy cross-channel CPC constraints.
\end{itemize}
The hyperparameters of the above methods follow the default settings in previous works. Table~\ref{tab:para} shows the main hyperparameters of HMMCB, and the remaining hyperparameters are discussed in Appendix $ \text{\uppercase\expandafter{\romannumeral3}}$.

\begin{table}[t]
\centering
\caption{Configurations and perparameters in HMMCB.}
\resizebox{\columnwidth}{!}{
\begin{tabular}{lccc}
\hline
\multicolumn{4}{c}{Parameters}                                                                                                       \\ \hline
\multicolumn{1}{c}{} & \multicolumn{1}{l}{top-level strategy} & \multicolumn{1}{l}{bottom-level strategy} & \multicolumn{1}{l}{CMCK} \\
batch size           & 7084                                   & 1024                                      & 2048                     \\
learing rates        & 1e-5                                   & 1e-5                                      & 2e-4                     \\
discounted factors   & 0.99                                   & 0.99                                      & -                        \\
decision interval    & 1 day                                  & each request                              & -                        \\
repetion times       & 25                                     & 8                                         & -                        \\
action range         & {[}0,1{]}                              & {[}0.5,1.5{]}                             & -                        \\ \hline
\end{tabular}
}
\label{tab:para}
\end{table}

\subsubsection{Evaluation metrics} To assess the performance of HMMCB, we incorporate four metrics for a comprehensive evaluation of multi-channel bidding scenarios, including total \textit{impressions} (\textbf{IMPR}), total \textit{clicks} (\textbf{CLICKS}), average \textit{cost per click} (\textbf{CPC}) and average \textit{return on investment} (\textbf{ROI}) across all advertisers.
Simply put, IMPR, representing the number of times an advertisement is displayed, indicates the level of activity among customers for that ad slot.
CLICKS represents the number of times that customers click on an advertisement, reflecting the level of interest in the advertised product.
CPC is the cost paid by advertisers for each click on an ad, and a small CPC means a low advertising cost for advertisers.
ROI represents the percentage of value that advertisers recoup from ad placements relative to the cost of advertising, and a large ROI represents a higher return on investment.
Note that ROI is the important comparison metric in all experiments.

In offline evaluation, in order to accurately evaluate the effectiveness of various methods and mitigate the influence of specific application scenarios, we employ a normalized scoring approach based on the statistical results obtained from the log data and use the offline evaluation system for testing. 
In online evaluation,  we use the A/B testing~\cite{DBLP:ZhangTYATXLX21} for each metric.
Specifically, we employ 1$\%$ of the whole customer traffic to independently conduct a two-week online experiment across four ad channels.
Note that each channel has an average of approximately 35,000 ad requests daily.

\begin{figure*}[t]
  \centering
  \subfloat{
    \label{fig:subfig_a}
    \includegraphics[scale=0.55]{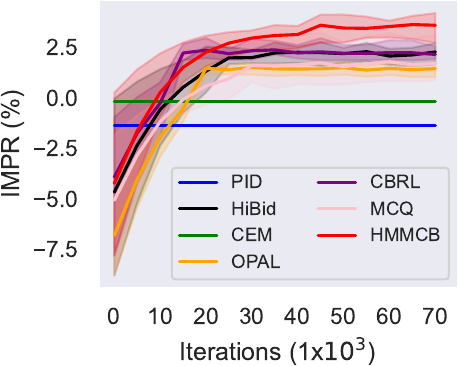}
  }
  \subfloat{
    \label{fig:subfig_b}
    \includegraphics[scale=0.55]{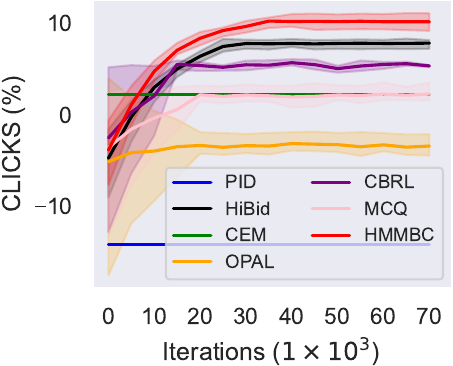}
  }
  \subfloat{
    \label{fig:subfig_d}
    \includegraphics[scale=0.55]{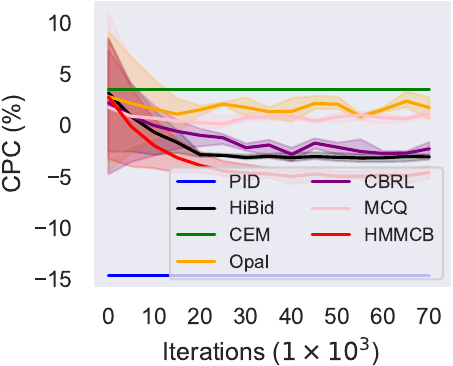}
  }
  \subfloat{
    \label{fig:subfig_c}
    \includegraphics[scale=0.55]{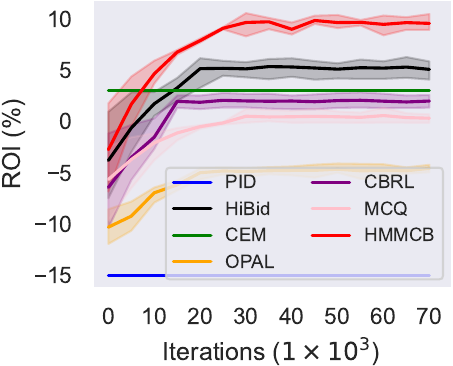}
  }
  \caption{Offline convergence process of the six methods. The vertical axis of each subplot represents different evaluation metrics, while the horizontal axis denotes the number of training steps. Each method is executed with 10 randomly selected seeds.}
  \label{fig:four_subfig}
\end{figure*}

\subsection{Performance Comparison}

\subsubsection{Offline performance comparison} 
In the offline setting, HMMCB is compared with PID, CEM, OPAL, CBRL and MCQ. Results in Table~\ref{tab:ofline_comp} indicate that HMMCB excels in maximizing both the ROI and IMPR, and effectively reduces the CPC, thereby enhancing the overall cost-effectiveness of the system. 
Although CBRL can progressively adapt its bidding strategy to meet constraints through the learning process, its performance still falls short of HMMCB, which can be attributed to the inability to allocate the budget accurately.
MCQ uses a conservative strategy to prevent overestimation of errors in out-of-distribution actions, but it suppresses the generalization of the value function and hinders performance improvement, resulting in a 1.91$\%$ increase in CPC.
OPAL demonstrates a 1.20$\%$ improvement in IMPR, but a decrease in  CLICKS, which can be attributed to the fact that the budget allocation strategy learned through supervised learning lacks appropriate adjustments in unfavorable situations. 
Due to limitations in policy representation capability, CEM struggles to effectively accommodate advertisers’ requirements, resulting in a significant increase in CPC to 3.59$\%$. 
While dynamically adjusting bidding prices based on the current CPC, PID fails to consider the variability in average CPC across different channels, leading to a 14.91$\%$ improvement in CPC, but decrease in all other metrics.
HiBid introduces auxiliary losses to prevent channel-specific over-allocation and uses $\lambda$-generalization to adapt to budget changes, but it ignores the internal connections between individual channels, resulting in only a 5.35$\%$ ROI improvement.

Compared with these models, HMMCB uses a diffusion strategy with CPC constraints to dynamically allocate budgets according to the characteristics of each channel, and uses the state action decoupled actor-critic method to avoid overestimation of out-of-distribution actions.
In addition, HMMCB is trained using the CMCK and central value function, allowing it not only to consider the specific attributes of each channel but also to coordinate all channels to achieve a global optimal performance. 
As a result, HMMCB achieves the highest ROI while ensuring superiority in other metrics.
A comprehensive depiction of the training convergence process is presented in Fig.~\ref{fig:four_subfig}.
Compared with other single-channel methods, HMMCB requires simultaneous learning of the characteristics of different channels to provide the best overall bidding solution, hence it requires multiple iterations to reach convergence.

\begin{table}[t]
\caption{Results of offline evaluation in terms of a relative improvement compared to the original method applied in the Meituan bidding system. 
$*$ denotes the primary metric  for comparison.
}
\begin{center}
\begin{tabular}{l|cccc}
\hline
methods & IMPR    & CLICKS  & CPC   & ROI*      \\ \hline
PID     & -1.30\% & -15\%   & \textbf{-14.91\%} & -15.0\% \\
CEM     & -0.21\% & 3.02\%  & 3.59\%  & 2.45\%   \\
OPAL    & 1.20\%  & -4.12\% & 3.41\% & -5.01\%   \\
CBRL    & 2.11\%  & 4.50\%  & -3.12\%  & 2.45\%  \\
MCQ     & 2.05\%  & 2.90\%  & 1.91\%  & 0.35\%   \\
HiBid     & 2.15\%  & 7.93\%  & -2.91\%  & 5.35\%   \\
HMMCB   & \textbf{3.97\%}  & \textbf{9.35}\%  & -3.40\%  & \textbf{9.80\%}   \\ \hline
\end{tabular}
\label{tab:ofline_comp}
\end{center}
\end{table}

\begin{figure}[t]
  \centering
  \subfloat[\scriptsize{Online A/B Testing on the Meituan online advertising bidding system. Each boxplot shows the average and median results of 30 independent repeated runs.}]{
    \includegraphics[scale=0.45]{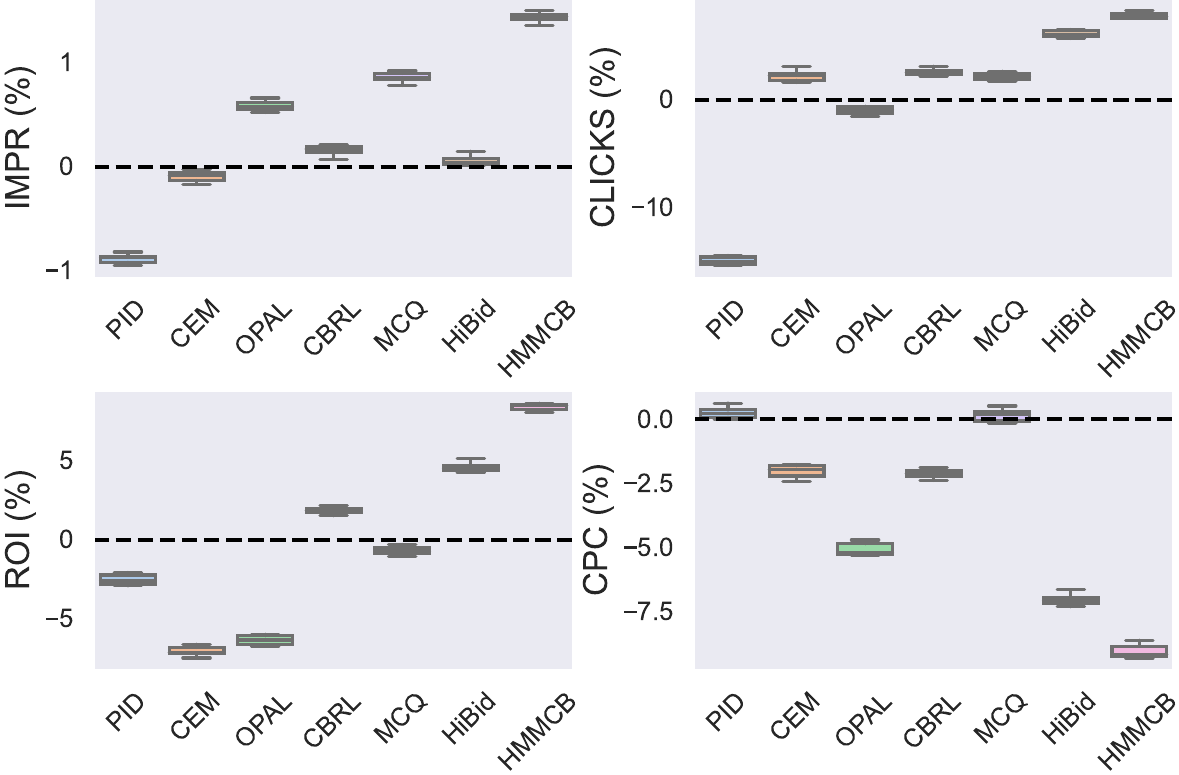}
    \label{fig:online_comp}
  }
  \\
  \subfloat[\scriptsize{99.9$\%$ request completion time for six methods. The system's response warning time is 50 milliseconds.}]{
    \includegraphics[scale=0.85]{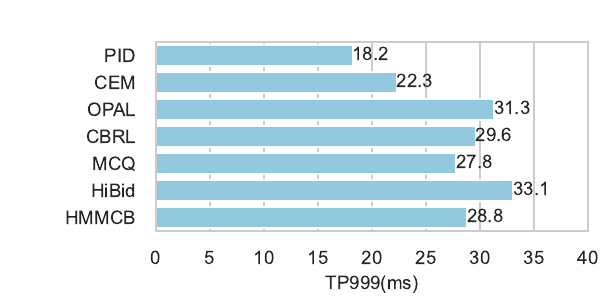}
    \label{fig:online_time}
  }
  \caption{Results of online experiments with seven methods}
  \label{fig:online_all}
\end{figure}

\subsubsection{Online A/B testing on the Meituan advertising platform}
\label{sec:online testing}
HMMCB and all other five baselines are validated using the Meituan advertising platform across four channels: feed ad, search ad, brand ad, and recommendation ad, with each method running a total test time of two weeks. 
As depicted in Fig.~\ref{fig:online_comp}, HMMCB surpasses all other baseline methods in all metrics, achieving an increase of at least 1.15$\%$ in IMPR, 8.30$\%$ in CLICKS, 8.65$\%$ in ROI, and a decrease of 8.83$\%$ in CPC. 
Fig.~\ref{fig:online_time} displays the TP999 (i.e., completion time for 99.9$\%$ requests) for each model.
While the HMMCB method takes into account the characteristics of different channels during training, each channel's strategy is executed independently during the evaluation, therefore it can still respond to requests in a relatively short amount of time.

In addition, we show that HMMCB can continually improve its performance by swiftly adapting to the evolving environment. To this end, we conduct the following iterated training process: (1) Daily bidding logs are automatically processed and stored in the server; (2) After one cycle, HMMCB is fine-tuned using the aforementioned offline data; and (3) After offline training, the new HMMCB model is synchronized to the RTB system for online bidding services.
We train four versions of HMMCB with a 5-day fine-tuning cycle and still use A/B testing for online experimental evaluation.
Table~\ref{tab:fine_online} indicates that HMMCB can achieve improvements in each evaluation metric after continuous fine-tuning and updating.

\begin{table}[t]
\caption{Fine-tune the test results of the model. HMMCB0 represents the model that has not been fine-tuned, and HMMCB1$-$HMMCB3 represent the models that have been fine-tuned based on the previous cycle model. }
\centering
\begin{tabular}{l|llll}
\hline
                                                                          & IMPR                                                                      & CLICKS                                                                    & CPC                                                                       & ROI*                                                                           \\ \hline
\begin{tabular}[c]{@{}l@{}}HMMCB0\\ HMMCB1\\ HMMCB2\\ HMMCB3\end{tabular} & \begin{tabular}[c]{@{}l@{}}1.15\%\\ 2.97\%\\ 3.99\%\\ \textbf{4.11}\%\end{tabular} & \begin{tabular}[c]{@{}l@{}}8.30\%\\ 8.36\%\\ 8.44\%\\ \textbf{8.62}\%\end{tabular} & \begin{tabular}[c]{@{}l@{}}-8.83\%\\ -8.88\%\\ -8.90\%\\ \textbf{-8.95}\%\end{tabular} & \begin{tabular}[c]{@{}l@{}}8.65\%\\ 9.35\%\\ 9.68\%\\ \textbf{9.89}\%\end{tabular} \\ \hline
\end{tabular}
\label{tab:fine_online}
\end{table}

\subsubsection{Effects of shared knowledge learning}
We adjust the data portion of the brand ad channel in the offline dataset against the other three channels as 1:50:50:50  and use this dataset to train PID, CEM, MCQ, CBRL, OPAL, HiBid, and HMMCB. 
Table~\ref{tab:meta_cmck} shows the impact of data scarcity on strategy learning.
The bidding performance of PID, CEM, MCQ, CBRL, OPAL, and HiBid algorithms drops significantly due to insufficient data for learning the characteristics of the brand channel.
HMMCB achieves superior results by leveraging CMCK to learn the shared knowledge of different channels and using this shared knowledge to enhance policy learning for the characteristics of the brand ad channel.

\begin{table}[t]
\centering
\caption{The data in the table is an offline evaluation of the brand ad channel. }
\begin{tabular}{lllll}
\hline
                                                                              & IMPR                                                                                           & CLICKS                                                                                           & CPC                                                                                              & ROI*                                                                                          \\ \hline
\begin{tabular}[c]{@{}l@{}}PID\\ CEM\\ OPAL\\ CBRL\\ MCQ\\ HiBid\\ HMMCB\end{tabular} & \begin{tabular}[c]{@{}l@{}}0.12\%\\ -1.49\%\\ -8.33\%\\ 2.13\%\\ -3.42\%\\ -1.31\% \\ \textbf{1.73}\%\end{tabular} & \begin{tabular}[c]{@{}l@{}}-6.53\%\\ -2.36\%\\ -1.49\%\\ -1.45\%\\ -7.83\%\\ 0.23\%\\ \textbf{6.43}\%\end{tabular} & \begin{tabular}[c]{@{}l@{}}0.95\%\\ 0.47\%\\ 2.15\%\\ 0.56\%\\ 2.63\%\\ -1.39\%\\ \textbf{-5.39}\%\end{tabular} & \begin{tabular}[c]{@{}l@{}}-1.23\%\\  0.31\%\\ -8.33\%\\ -8.76\%\\ -2.34\%\\ -2.78\%\\ \textbf{2.01}\%\end{tabular} \\ \hline
\end{tabular}
\label{tab:meta_cmck}
\end{table}

\subsection{Further Study}
\begin{figure}[t]
  \centering
  \subfloat[]{
    \label{fig:sate_action_roi}
    \includegraphics[scale=0.55]{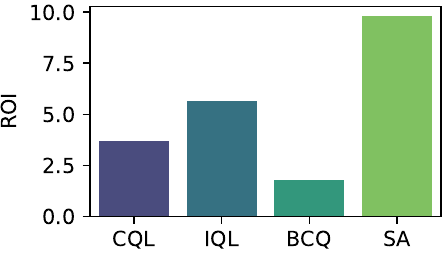}
  }
  \subfloat[]{
    \label{fig:cmck_roi}
    \includegraphics[scale=0.55]{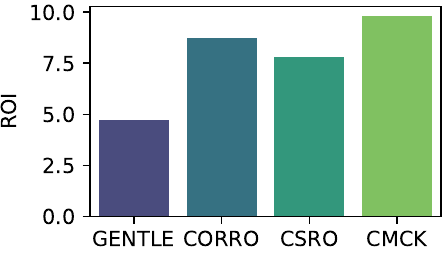}
  }
  \caption{Comparison results of HMMBC with existing methods. The experimental results are based on running the experiment 5 times with a random seed, and the y-axis represents the ROI. SA denotes the state-action decoupled actor-critic method.
}
  \label{fig:5_subfig}
\end{figure}

In this section, we explore the comparison of the state-action decoupled actor-critic method and the context-based knowledge sharing approach (CMCK) with several existing works.
Specifically, we replace the state-action decoupled actor-critic method with mainstream offline RL methods CQL~\cite{DBLP:LyuMLL22}, IQL~\cite{DBLP:KostrikovNL22}, and BCQ~\cite{DBLP:FujimotoMP19}, respectively.
Similarly, we substitute the context-based knowledge-sharing approach with offline meta-RL methods, including GENTLE~\cite{zhou2024generalizable}, CORRO~\cite{DBLP:YuanL22}, and CSRO~\cite{DBLP:GaoZGWYPLCDHGLC23}.
Fig.~\ref{fig:sate_action_roi} shows that the state-action decoupled actor-critic method consistently outperforms the CQL, IQL, and BCQ methods.
CQL and BCQ prioritize learning policies that are very similar to behavioral policies, which limits their ability to leverage non-expert data to enhance policy performance.
IQL reduces extrapolation errors in the policy improvement process by avoiding the direct evaluation of actions that are not present in the dataset.
However, IQL employs advantage-weighted behavior cloning to learn a policy, which may be inadequate for capturing the optimal policy distribution in a bidding dataset containing a significant proportion of non-expert data. 
Fig.~\ref{fig:cmck_roi} shows the comparison results between CMCK and Meta-RL methods, indicating that CMCK achieves the highest ROI, while GENTLE results in the lowest ROI. GENTLE increases channel data by reconstructing state transitions and rewards, which may bring larger distribution errors, causing the learned policy to deviate from the actual data distribution.

\begin{table}[]
\centering
\caption{Comparison Results of Offline Multi-Agent Methods
}
\begin{tabular}{lllll}
\hline
                                                                          & IMPR                                                                                  & CLICKS                                                                               & CPC                                                                                  & ROI*                                                                                  \\ \hline
\begin{tabular}[c]{@{}l@{}}MAAB\\ ICQ\\ OMAR\\ OMIGA\\ HMMCB\end{tabular} & \begin{tabular}[c]{@{}l@{}}0.63\%\\ 0.13\%\\ -1.42\%\\ -0.31\% \\ \textbf{3.97}\%\end{tabular} & \begin{tabular}[c]{@{}l@{}}1.22\%\\ -5.45\%\\ -6.23\%\\ 0.33\%\\ \textbf{9.53}\%\end{tabular} & \begin{tabular}[c]{@{}l@{}}2.98\%\\ 5.32\%\\ 3.63\%\\ -0.59\%\\ \textbf{-3.40}\%\end{tabular} & \begin{tabular}[c]{@{}l@{}}0.78\%\\ -5.76\%\\ -1.94\%\\ -1.78\%\\ \textbf{9.80}\%\end{tabular} \\ \hline
\end{tabular}
\label{tab:multi_comp}
\end{table}

Additionally, HMMCB is compared with existing multi-agent methods to highlight its practicality in multi-channel bidding scenarios.
Specifically, we compare the bidding multi-agent algorithm MAAB~\cite{DBLP:WenXZZWLRXTYXWC22}, which uses the temperature-regularized credit assignment to handle bidding relationships between different channels.
The original MAAB algorithm does not meet the requirements of existing bidding environments, as it does not consider channel budgets and is designed for settings where channels exhibit cooperative-competitive relationships.
We modify the MAAB to align with the cooperative environment settings and allocate budgets proportionally according to the historical revenue ratios of each channel.
Moreover, we also consider representative offline MARL algorithms, ICQ~\cite{DBLP:YangMLZZHYZ21}, OMAR~\cite{DBLP:PanH0X22}, and OMIGA~\cite{DBLP:WangXZZ23}, with the core idea of replacing global regularization with local regularization to learn the optimal joint policy.
The results in Table~\ref{tab:multi_comp} indicate that HMMCB consistently outperforms the MAAB, ICQ, OMAR, and OMIGA methods.
This superior performance arises not only from HMMCB's effective resolution of the out-of-distribution (OOD) problem but also from its dynamic budget allocation and efficient handling of different channel characteristics. 
In contrast, the ICQ, OMAR and OMIGA algorithms focus on solving the offline multi-agent OOD problem, which makes them unable to obtain satisfactory performance in complex and stochastic bidding scenarios because different channel budgets and customer traffic can affect bidding.

\subsection{Ablation Study}
In this section, we modify each component within HMMCB to demonstrate their individual effectiveness, with the results given in Table~\ref{tab:ablation_offline}.
First, we evaluate the influence of top-level strategies by comparing bidding policies with and without CPC constraints. 
Without a CPC constraint, HMMCB can disregard the overall CPC to maximize clicks, leading to a 6.25$\%$ surge in clicks while also causing a 1.34$\%$ rise in CPC.
Second, we examine the influence of policy regularization capability, specifically focusing on whether employing diffusion model as a top-level strategy can enhance the effectiveness of our bidding model.
Results show employing a traditional MLP as a top-level strategy degrades the overall model performance, due to its incapability in accurately capturing the data distribution across multiple channels.
Finally, we explore the impact of CMCK and central value function on the final performance. 
In scenarios where the state policy is not learned using CMCK, independently trained state policies struggle to effectively generalize the optimal next state for each channel. 
Hence, it incurs a reduction of 1.34$\%$ in ROI and an increase of 3.25$\%$ in CPC. 
When each bottom-level strategy undergoes independent training without reliance on a central value function, there is a notable 7.43$\%$ decrease in ROI, despite a 9.54$\%$ increase in clicks. 
This discrepancy can be attributed to the fact that each policy solely focuses on maximizing its own returns without considering the interrelationships among different channels, leading to a local optimum in overall performance.

\begin{table}[t]
\caption{Ablation Study.}
\centering
\begin{tabular}{lcccc}
\hline
method                                                                                                                      & IMPR                                                                               & CLICKS                                                                              & CPC                                                                                   & ROI*                                                                                 \\ \hline
\begin{tabular}[c]{@{}l@{}}HMMC\\ without CPC\\ without Diffusion\\ without CMCK\\ without Central\end{tabular} & \begin{tabular}[c]{@{}c@{}}\textbf{3.97}\%\\ 1.27\%\\ 3.74\%\\ 3.03\%\\ 3.73\%\end{tabular} & \begin{tabular}[c]{@{}c@{}}9.35\%\\ 6.25\%\\ 0.36\%\\ 4.54\%\\ \textbf{9.54}\%\end{tabular} & \begin{tabular}[c]{@{}c@{}}\textbf{-3.40}\%\\ 1.34\%\\ 6.23\%\\ 3.25\%\\ 8.25\%\end{tabular} & \begin{tabular}[c]{@{}c@{}}\textbf{9.80}\%\\ 2.34\%\\ -2.34\%\\ -1.34\%\\ -7.34\%\end{tabular} \\ \hline
\end{tabular}
\label{tab:ablation_offline}
\end{table}

\begin{figure}[t]
    \centering
    \includegraphics[scale=0.55]{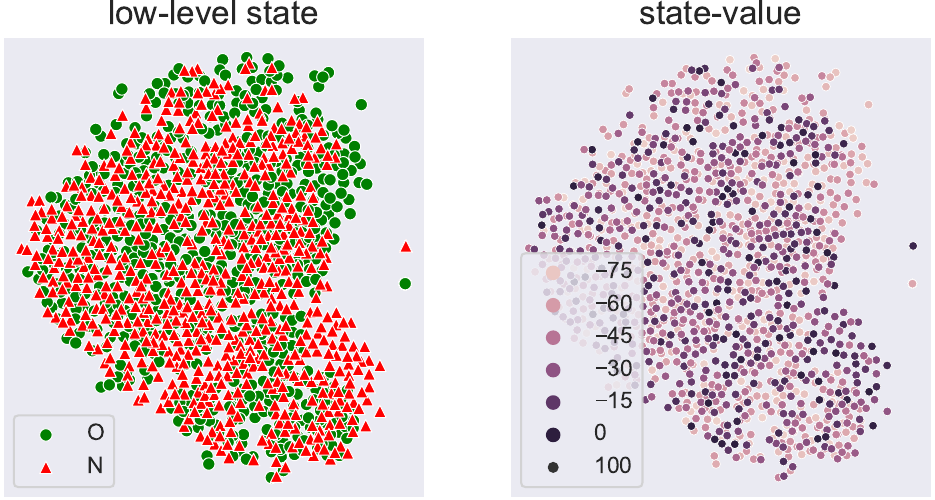}
    \caption{In the left, $\text{O}$ (green dot) represents the original state distribution, while $\text{N}$ (red $\triangle$) represents the states generated by the state policy. The right displays state reward values, with darker colors indicating higher values.}
    \label{fig:tsne_iamge}
\end{figure}

In addition, we use SNE~\cite{DBLP:HintonR02} to visualize the feature distribution of $\pi_s(s_t)$ and $s_{t+1}$ to illustrate how the state strategy guides the action strategy in bidding decisions.
The results are shown in Fig.~\ref{fig:tsne_iamge}, where the left image illustrates the distribution of the state, while the right image represents the reward value for each state.
We can see that the state distribution of $\pi_{s}$ follows the distribution of the original data and achieves higher returns, which is crucial in guiding the action policy to make better decisions through accumulating and propagating high return  through multiple time steps.

\section{Conclusion}

In this paper, we conduct in-depth analysis of the cross-channel bidding problem and propose a hierarchical optimization architecture HMMCB to solve it. 
HMMCB enables the dynamic allocation of channel budgets by implementing a CPC-constrained diffusion model, and uses a state-action decoupled actor-critic method and a context-based meta-channel knowledge learning method to utilize the cross-channel interrelationships for efficient bidding decision making within each channel.
Both offline and online experiments based on industry data from the Meituan ad bidding platform validate the superiority of HMMCB against some traditional or state-of-the-art methods.

\section*{Acknowledgments}
We gratefully acknowledge support from the National Natural Science Foundation of China (No. 62076259), the Fundamental and Applicational Research Funds of Guangdong province (No. 2023A1515012946), and the Fundamental Research Funds for the Central Universities-Sun Yat-sen University. This research was supported by Meituan.



 
%

\bibliographystyle{IEEEtran}
\bibliography{IEEEabrv}


 


\begin{IEEEbiography}[{\includegraphics[width=1in,height=1.25in,clip,keepaspectratio]{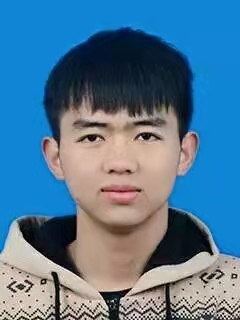}}]{Shenghong He}
is currently a PhD student under the  supervision of Prof. Chao Yu in the School of Computer Science and Engineering, Sun Yat-sen University, China. He has published  a few works at IEEE TDSC, and is now working on the problem of  offline learning and security with deep reinforcement learning.
\end{IEEEbiography}

\vspace{1pt}

\begin{IEEEbiography}[{\includegraphics[width=1in,height=1.25in,clip,keepaspectratio]{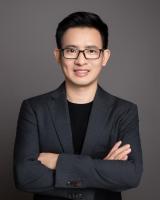}}]{Chao Yu}
 received the Ph.D. degree in computer science from the University of Wollongong, Australia,  in 2014. He is currently an Associate Professor with  the School of Computer Science and Technology,  Dalian University of Technology, Dalian, China,  and a Visiting Professor with the Department of  Computer Science, Hong Kong Baptist University,  Hong Kong. He has published more than 50 papers  in prestigious journals, such as the IEEE TNNLS,  IEEE TCYB, and IEEE TVT. His research interests  include multi-agent systems, reinforcement learning,  and their wide applications in autonomous driving, smart grid, robotic control,  and intelligent healthcare.
\end{IEEEbiography}

\begin{IEEEbiography}[{\includegraphics[width=1in,height=1.25in,clip,keepaspectratio]{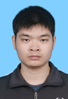}}]{Qian Lin}
, a Master's student in Sun Yat-sen University, China, supervised by Professor Chao Yu. He mainly focuses on reinforcement learning (RL), especially under various settings, including offline, safe, and multi-objective scenarios. He has published several research in top venues, including ICML, AAMAS and ICLR.
\end{IEEEbiography}

\begin{IEEEbiography}[{\includegraphics[width=1in,height=1.25in,clip,keepaspectratio]{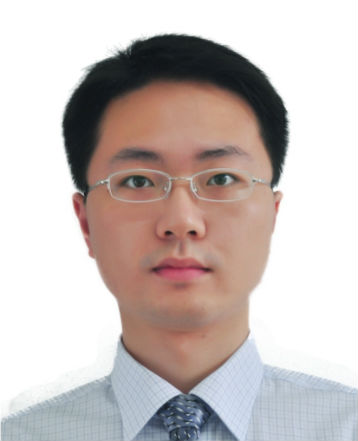}}]{Shangqin Mao}
received his Ph.D. degree from  Huazhong University of Science and Technology in  Hubei, China in 2013. He currently works as a data  scientist at Meituan. His current research interests  include auction theory and reinforcement learning.
\end{IEEEbiography}

\begin{IEEEbiography}[{\includegraphics[width=1in,height=1.25in,clip,keepaspectratio]{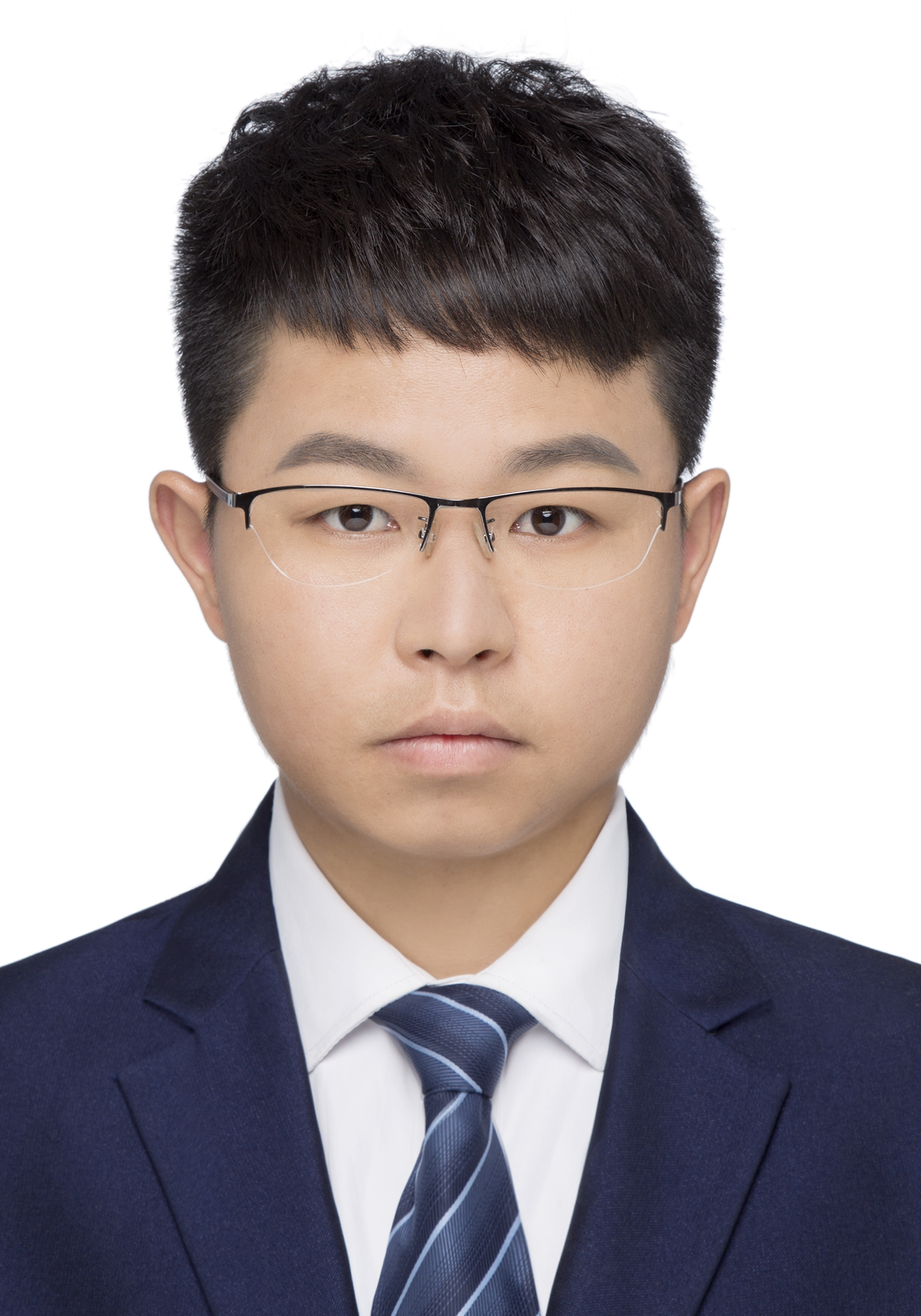}}]{Bo Tang}
is currently a PhD student under the supervision of Prof. Jingrun Chen in the School of Artificial Intelligence and Data Science, University of Science and Technology of China. He has published several works at major conferences such as ICML, NeurIPS, WWW, and CIKM, and is now focusing on the problem of advertising bidding using deep reinforcement learning.
\end{IEEEbiography}

\begin{IEEEbiography}[{\includegraphics[width=1in,height=1.25in,clip,keepaspectratio]{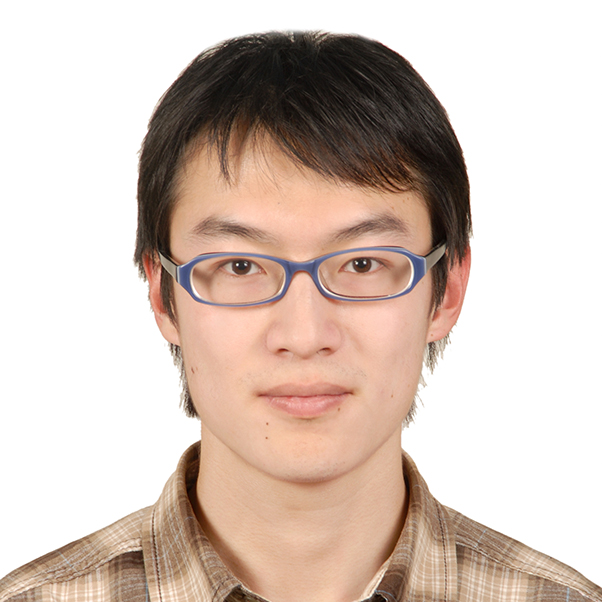}}]{Qianlong Xie}
received his M.S. degree from Beijing University of Posts and Telecommunications in  Beijing, China in 2013. He is currently employed  as a machine learning scientist at Meituan. He has  published several works at prestigious conferences  such as ICML, WWW, and DASFAA. His primary  research interests include image generation, machine  learning, and data mining.
\end{IEEEbiography}

\begin{IEEEbiography}[{\includegraphics[width=1in,height=1.25in,clip,keepaspectratio]{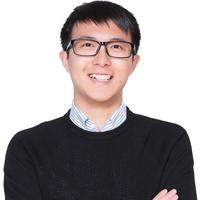}}]{Xingxing Wang}
earned his M.S. degree from the  Computer Network Information Center at the Chinese Academy of Sciences in Beijing, China in 2013.  He is currently employed as a machine learning  scientist at Meituan. He has published several works  at prestigious conferences such as ICML, WWW,  AAAI, CIKM, and DASFAA. His primary research  interests include recommender systems, machine  learning, and data mining.
\end{IEEEbiography}



\vfill

\end{document}